\documentclass{article}
\pdfoutput=1
\usepackage[final]{neurips_2020}
\usepackage[utf8]{inputenc} 
\usepackage[T1]{fontenc}    
\usepackage{hyperref}       
\usepackage{url}            
\usepackage{booktabs}       
\usepackage{amsfonts}       
\usepackage{nicefrac}       
\usepackage{microtype}      
\usepackage[english]{babel}
\usepackage{graphicx}
\usepackage{float}
\usepackage{amsfonts,amssymb,amsmath,amsthm} 
\usepackage{todonotes}
\usepackage{pgfplots,subcaption}

\pgfplotsset{colormap/viridis}

\usepackage{caption}
\usepackage[shortlabels]{enumitem}
\usepackage{booktabs}

\usetikzlibrary{backgrounds,automata}

\renewcommand{\P}{{\mathbb{P}}}
\newcommand{\E}{{\mathbb{E}}}

\newtheorem{theorem}{Theorem}[section]

\newtheorem{corollary}[theorem]{Corollary}
\newtheorem{proposition}[theorem]{Proposition}

\theoremstyle{definition}
\newtheorem{definition}[theorem]{Definition}

\theoremstyle{remark}
\newtheorem{remark}[theorem]{Remark}

\title{Towards an intrinsic definition of robustness\\for a classifier}

\author{Théo Giraudon$^1$, Vincent Gripon$^{1,2}$, Matthias Löwe$^3$, Franck Vermet$^4$\\
$^1$IMT Atlantique, $^2$Université Côte d'Azur, \\$^3$University of Münster, $^4$Université de Bretagne Occidentale}
\date{March 2020}

\begin{document}

\maketitle

\begin{abstract}
The robustness of classifiers has become a question of paramount importance in the past few years. Indeed, it has been shown that state-of-the-art deep learning architectures can easily be fooled with imperceptible changes to their inputs. Therefore, finding good measures of robustness of a trained classifier is a key issue in the field. In this paper, we point out that averaging the radius of robustness of samples in a validation set is a statistically weak measure. We propose instead to weight the importance of samples depending on their difficulty. We motivate the proposed score by a theoretical case study using logistic regression, where we show that the proposed score is independent of the choice of the samples it is evaluated upon. We also empirically demonstrate the ability of the proposed score to measure robustness of classifiers with little dependence on the choice of samples in more complex settings, including deep convolutional neural networks and real datasets.
\end{abstract}

\section{Introduction}

Deep Learning is the golden standard for many challenges in the field of machine learning. Thanks to a large number of tunable parameters, these architectures are able to absorb subtle dependencies from large datasets and then to generalize decisions to previously unseen inputs. Among other examples, Deep Learning is state-of-the-art in classification in vision~\cite{he2015delving}, playing complex abstract games~\cite{mnih2013playing}, processing natural language~\cite{young2018recent}, or decoding the brain activity~\cite{varoquaux2019predictive}.

However, despite achieving outstanding results, as typically measured as the accuracy on a validation or test set, Deep Learning architectures are likely to fail at correctly classifying corrupted inputs. This limitation has first been shown in~\cite{szegedy2013intriguing}, where the authors showed it is possible to fool the system decision using a humanly imperceptible additive noise. This finding opened the way to a lot of contributions where increasingly efficient attacks have been introduced~\cite{moosavi2016deepfool}. If such attacks could be considered artificial, in the sense that they require access to the network function, recent works ~\cite{chen2017zoo,hendrycks2019benchmarking} have shown that some black-box corruptions (where inputs are modified agnostically of the network function) are likely to dramatically lower the accuracy of the system as well.

In the recent years, there have been an increasing number of works proposing mitigation strategies to enhance the robustness of trained architectures. By robustness, most authors think of the ability of the network function to maintain a good decision for a certain radius around the training or test samples~\cite{ruan2018global}. Examples of such strategies are listed in Section~\ref{rw}.

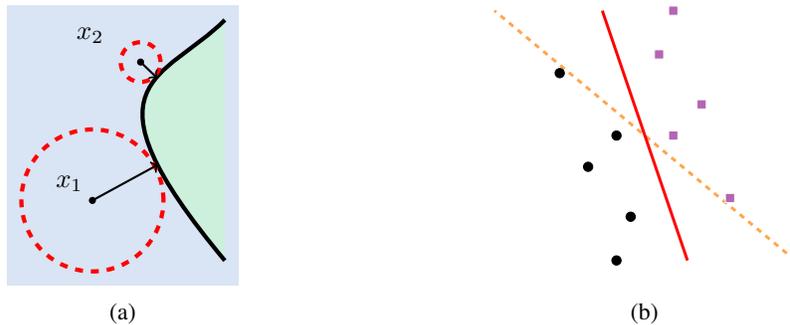
\begin{figure}[h]\centering
	\begin{subfigure}[t]{0.49\textwidth} \centering
\begin{tikzpicture}[scale=0.8, background rectangle/.style={fill=blue!70!green!15}, show background rectangle]
\draw[black,ultra thick, fill=blue!30!green!20](4,0) .. controls (
1.5,3) and (3,3)  ..  (4,4);
\draw [red, ultra thick, dashed] (1.8,1) circle[radius=1.18];
\draw [red, ultra thick, dashed] (2.6, 3.3) circle[radius=0.33];
\foreach \Point/\RobustPoint in {(1.8,1)/x_1}
\draw[fill=black] \Point circle (0.05) node[above left] {$\RobustPoint$};
\foreach \Point/\NonRobustPoint in {(2.6, 3.3)/}
\draw[fill=black] \Point circle (0.05) node[label={[xshift=-19, yshift=0]$x_2$}] {$\NonRobustPoint$};
 \coordinate (A1) at (1.8,1) ;
\coordinate (B1) at (2.6, 3.3) ;
\coordinate (A2) at (2.9,1.6) ;
\coordinate (B2) at (2.88, 3.03) ;
\draw[->, thick] (A1) -- (A2);
\draw[->, thick] (B1) -- (B2);
\end{tikzpicture}
		\subcaption{}
		\label{radius_robustness}
	\end{subfigure}
	\begin{subfigure}[t]{0.49\textwidth} \centering
		
		\begin{tikzpicture}[scale = 0.7]
		\begin{axis}[
		scatter/classes={%
			a={mark size=2.5pt},%
			b={mark size=2.5pt, mark=square*, draw=white, fill=violet!60}},
		hide axis,          
		mark=none]
		\addplot[scatter,only marks,%
		scatter src=explicit symbolic]%
		table[meta=label] {
			x     y      label	
			0.9   0       a 
			0.9   -1      a 
			0.7   0.5     a 
			0.95  -0.65   a 
			0.8   -0.25   a 
			1.1   0       b 
			1.1   1       b
			1.3   -0.5    b
			1.05  0.65    b
			1.2   0.25    b
		};
		\addplot[color=orange!70,mark=, dashed, ultra thick] coordinates { (1.53,-1) (0.47,1) };
		\addplot[color=red,mark=, ultra thick] coordinates { (1.15,-1) (0.85,1) };
		\end{axis}
		\end{tikzpicture}
		\subcaption{}
		\label{two_classifiers}
	\end{subfigure}
	\caption{Figure (\ref{radius_robustness}) represents the radii of robustness of two points $x_1$ and $x_2$ chosen in the same class region. We observe that the maximum radius of robustness for $x_1$ is much larger than that for $x_2$, as a consequence of their respective distances to the boundary. This drawing emphasizes the fact that \emph{not} all samples should be expected to have the same radius of robustness.
	\\
	Figure (\ref{two_classifiers}) shows two classes separated by two linear classifiers. When estimating robustness of the classifier as the mean of the radius of each sample, we obtain that the orange dotted line is the preferred one. When we account for the difficulty of samples, using our definition~\ref{def:robusnessours}, we obtain that the red solid line is the most robust. The latter is indeed more robust, as it yields a larger margin.}
	\label{bigfigure}
\end{figure}

Robustness is intrinsically tied with generalization. Indeed, an ideal network function that would correctly predict the class of any input would by definition be robust. Interestingly, let us point out that this statement implies that even in the case of a perfect classifier, robustness cannot be thought of uniformly for any input in a class domain. As a matter of fact, a classifier defines a partition of the input domain in class regions. Hence some inputs are arbitrarily close to the boundary. It cannot be expected that such inputs have a large radius of robustness. On the contrary, inputs chosen at the center of class regions are likely to yield large such radii. This phenomenon is depicted in Figure~\ref{bigfigure}.

In this work, we introduce a new definition of robustness which aims at simplifying the comparison of the intrinsic robustness of classifiers, by reducing the impact of the choice of samples it is estimated on. More precisely, we have the following claims:
\begin{enumerate}
    \item We introduce a new definition of robustness that is proved independent of the choice of the samples it is estimated in the case of the logistic regression,
    \item We perform experiments using challenging vision datasets and corruption benchmarks to validate the interest of the proposed definition in a practical scenario.
\end{enumerate}

\begin{figure}[h]
     \centering
     \begin{subfigure}[b]{0.22\textwidth}
         \centering
         \includegraphics[width=\textwidth]{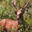}
         \caption{Loss = 0.005}
         \label{deer1}
     \end{subfigure}
     \hfill
     \begin{subfigure}[b]{0.22\textwidth}
         \centering
         \includegraphics[width=\textwidth]{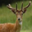}
         \caption{Loss = 0.006}
         \label{deer2}
     \end{subfigure}
     \hfill
     \begin{subfigure}[b]{0.22\textwidth}
         \centering
         \includegraphics[width=\textwidth]{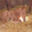}
         \caption{Loss = 2.584}
         \label{horse}
     \end{subfigure}
     \hfill
     \begin{subfigure}[b]{0.22\textwidth}
         \centering
         \includegraphics[width=\textwidth]{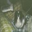}
         \caption{Loss = 2.655}
         \label{dog}
     \end{subfigure}
        \caption{These pictures come from the CIFAR-10 dataset. The losses (cross-entropy) that are indicated under them are computed using a trained ResNet18. (\ref{deer1}) and (\ref{deer2}) have the lowest losses of the validation set, meanwhile (\ref{horse}) and (\ref{dog}) have the highest ones. As can be seen, the highest loss correspond to samples that are more difficult to classify even for the human eye.}
        \label{pictureswithdifferentlosses}
\end{figure}

\section{Related Work}
\label{rw}

\textbf{Adversarial attacks}

Adversarial attacks have been first put into light by \cite{szegedy2013intriguing}. They consist of deformations with minimal norm, often imperceptible to the human eye, that causes the network to change its prediction with high confidence. To create these deformations, a complete knowledge about the decision function of the classifier, in particular its gradient, is needed. Numerous techniques have been created in order to find the smallest attack, like the Fast Gradient Sign Method \cite{goodfellow2014explaining}, the Projected Gradient Descent \cite{madry2017towards} or the Deep Fool algorithm \cite{moosavi2016deepfool} among many others. The existence of these attacks shows that neural networks of any kind can be sensible to small changes in an input, which causes important security problems, for instance in the context of autonomous driving \cite{eykholt2018robust}. 

\textbf{Black-box corruptions}

One could argue that these deformations need perfect knowledge about the network function and hence do not appear in practical life. However, \cite{hendrycks2019benchmarking} suggested black-box (i.e. agnostic of the network function) corruptions that considerably damage predictions of state-of-art neural networks on the CIFAR-10 and ImageNet datasets. These corruptions imitate natural effects like rain, fog, JPEG compression, brightness etc. Of course, they are no longer imperceptible, but they still are of interest because much more bound to happen in real life than adversarial attacks. In our work, we will take into account mainly adversarial attacks but also black-box corruptions in the appendix.

\textbf{Defence against the attacks}

In order to reduce the efficiency of these attacks, several options have been studied. One of the most popular is adversarial training \cite{goodfellow2014explaining,tramer2017ensemble,ganin2016domain}: the idea is to find adversarial attacks for the elements of the training set and to add them to it. It seems like this modified training also makes the network more robust to unknown types of deformations and not only to those encountered during the training. Apart from acting on the training set, another idea is to change the decision process, for instance through randomized smoothing \cite{cohen2019certified,salman2019provably} where the new prediction for an input is defined as the most likely prediction by the base classifier on random Gaussian corruptions of this input. This technique smooths the decision boundary and increases the distance of the inputs to it, thus increasing robustness. Finally, some authors choose to work on the structure of the network function by forcing its Lipschitz constant to be smaller than 1, either softly as it is the case in Parseval networks \cite{cisse2017parseval} or hardly~\cite{qian2018l2}. We will study the impact of some of these strategies on our proposed robustness score.

\textbf{Certification of robustness}

Instead of trying to make neural networks more robust to attacks that become increasingly more efficient, one could want to give guarantees about the robustness of a classifier. Some authors \cite{raghunathan2018certified} followed this path and suggested ways to find the maximum safety radius of a classifier on a dataset, that is the maximum $r\geq0$ such that any well-classified input is still well classified after any deformation of norm less than $r$. Some took a probabilistic approach \cite{weng2018proven} and consider the maximum $r\geq0$ such that there exists, with probability less than $\alpha$, an adversarial attack of norm less than $r$. We will show, both theoretically and empirically, that our certification is more robust that the classical deterministic one because way less dependant of the samples the network it is evaluated upon. 

\section{Methodology}

In this section we will present our theoretical approach to robustness. 

\subsection{The golden score of robustness}
\label{golden_score_section}
An ideal score of robustness $R_*(f, \mathcal{D})$ for a trained classifier $f$ on a dataset $\mathcal{D}$ would be one satisfying the followings properties:
\begin{enumerate}[(I)]
    \item (Monotony) for two classifiers $f_1$, $f_2$ with the same test accuracy on $\mathcal{D}$, if all well-classified points are further from the decision boundary of $f_1$ than from the one defined by $f_2$, then $R_*(f_1, \mathcal{D}) \geq R_*(f_2, \mathcal{D})$.
    \item (Separability) if all well-classified points from $\mathcal{D}$ are at distance 0 from the decision boundary defined by $f$, then $R_*(f, \mathcal{D})=0$.
    \item (Subset Independence) $R_*(f, \mathcal{D})$ does not depend on the samples from $\mathcal{D}$ is it evaluated on.
\end{enumerate}

(III) is of fundamental importance because the main objective of testing robustness is to ensure that in the considered applications the classifier will be robust, and not only on the considered validation set.

\subsection{The problems of mean-case and worst-case robustness}

Let us suppose we have a labelled dataset $\mathcal{D} = \left(\mathbf{x}_i, y_i\right)_i$ composed of $|\mathcal{D}|$ elements where $\mathbf{x}_i \in \mathbb{R}^d$ and $y_i \in \{1,\dots,K\}$. We denote by $f$ a trained classifier that associates any valid input $\mathbf{x}$ with a decision $f(\mathbf{x}) \in \{1, \dots, K\}$.

The robustness of $f$ is usually defined using the concept of the radius of robustness:
\begin{definition}
The (maximum) radius of robustness of $(\mathbf{x},y)$ with respect to the classifier $f$ is defined to be the largest $r\geq 0$ such that $$ \forall \mathbf{n}\in\mathbb{R}^d ~\text{with}~ \|\mathbf{n}\|_q < r, ~\text{we have}~ f(\mathbf{x}+\mathbf{n})=y, $$
where $\|\cdot\|_q$ is the $L^q$ norm.
We will denote this maximum radius of robustness by $r_f(\mathbf{x},y)$ or simply $r(\mathbf{x},y)$ if there is no ambiguity. Note that $r(\mathbf{x},y) = 0$ for misclassified pairs $(\mathbf{x},y)$.
\end{definition}

Then, most authors introduce a measure of robustness as a simple statistical value depending on $r(\mathcal{D}) = \{r(\mathbf{x}_i, y_i)_i\}$. Two very common examples are the mean-case robustness \cite{moosavi2016deepfool}, defined as:
\begin{equation}
    R_m(\mathcal{D}, f) = \frac{1}{|\mathcal{D}|} \sum_{i=1}^{|\mathcal{D}|}{r(\mathbf{x}_i, y_i)}, 
    \label{mean_score}
\end{equation}{}
and the worst-case robustness, defined as:
\begin{equation}
    R_w(\mathcal{D}, f) = \min_{i=1}^{|\mathcal{D}|}{r(\mathbf{x}_i, y_i)}. 
    \label{min_score}
\end{equation}

It is easy to check that both of these definitions satisfy properties (I) and (II), but not (III). In practice, robustness is measured using a validation set $\mathcal{D}$, that is supposed to be a proxy to real-world data. Obviously, the worst-case robustness is very sensitive to the choice of $\mathcal{D}$, as removing the worst case from it would likely change the measure of robustness. But even mean-case robustness is sensitive to how examples in $\mathcal{D}$ are drawn. Consider the toy illustration of Figure~\ref{bigfigure}a) for example, where we can see two samples that are likely to yield very distinct radii of robustness, even for the most robust of classifiers. As a consequence, sampling examples close to the boundary would cause a low value of robustness, whereas sampling far form the boundary would cause a larger value.

One could argue that since classifiers are usually compared over the same validation set, there is no major concern about this observation. Let us point out that considering the validation set is biased towards easy or hard samples, it is very likely that the difference measured between two classifiers in terms of robustness is a very weak statistical test.

As a matter of fact, wanting all samples to yield similar radii of robustness is not necessarily desirable in practice, as some of these samples might approach the boundary between class regions. Consider Figure~\ref{pictureswithdifferentlosses} where we depicted easy and hard samples to classify according to the corresponding loss of a trained classifier on the CIFAR-10 vision dataset. One would expect the easy samples to yield an important radius of robustness, whereas hard ones would yield smaller radii. Problematically, it is not known in advance precisely which samples are close to the boundary and which are far. In the next section, we show that the radii of robustness of samples can be directly linked to their corresponding loss for a trained classifier in the case of a logistic regression, so that it is possible to define a notion of robustness that is statistically robust to the sampling of the validation set $\mathcal{D}$ it is tested upon.

\subsection{Robustness and Loss}

Let us show how radius of robustness and loss can be dependent, in the illustrative case of a logistic regression. Consider $f$ to be a multinomial logistic regression with parameters $(\beta_j, \beta_j^0)_{j=1,..,K}$ where $\beta_j \in \mathbb{R}^d, \beta_j^0 \in \mathbb{R}$ and let $\mathcal{D} = \{(\mathbf{x}_i, y_i)_{1\leq i \leq |\mathcal{D}|}\}$ be a dataset where $\mathbf{x}_i\in\mathbb{R}^d$ and $y_i\in\{1,\dots,K\}$. Let $\ell(\mathbf{x},y)$ be the cross-entropy loss for a sample $(\mathbf{x},y)$, that is
\begin{equation}
	\ell(\mathbf{x}, k) = -\log \Bigg(\frac{\exp(\beta_k\cdot\mathbf{x}+\beta_k^0)}{\sum_{j=1}^K\exp(\beta_j\cdot\mathbf{x}+\beta_j^0)}\Bigg).
\end{equation}
We then have the following:

\begin{proposition}
\label{proposition_logistic}
For every well-classified input $\mathbf{x}$ belonging to class $k$, there exists $m\not=k$ such that
	\begin{equation}
		\frac{-1}{\|\beta_m-\beta_k\|}\log\left(\frac{1}{K-1}(\exp\ell(\mathbf{x},k)-1)\right) \leq r(\mathbf{x},k) \leq \frac{-1}{\|\beta_m-\beta_k\|}\log\left(\exp\ell(\mathbf{x},k)-1\right),
	\end{equation}
where $r$ is defined for $q=2$, i.e. the euclidean norm.
\end{proposition}
The proof of proposition (\ref{proposition_logistic}) can be found in the appendix. When $K=2$ we immediately have:

\begin{corollary}
\label{corollary_binomial_regression}
In the setting of binomial logistic regression, that is when the number of classes $K$ is equal to 2, we have 
\begin{equation}
    r(\mathbf{x},k) = \frac{-1}{\|\beta\|}\log\big(\exp\ell(\mathbf{x},k)-1\big),
\end{equation}
where we denoted $\beta := \beta_2-\beta_1$.
\end{corollary}

In the remaining of this work, we will denote $g(t)=-\log (\exp t - 1)$ and we will always consider $\ell(\mathbf{x},y)$ to be the cross-entropy loss of a sample $(\mathbf{x},y)$. In the context of binomial logistic regression, a direct consequence of Corollary \ref{corollary_binomial_regression} is that if we sample elements in $\mathcal{D}$ using a probability measure $\nu$ such that 
\begin{equation}
    \nu(\mathbf{x}, y) \propto \frac{1}{g(\ell(\mathbf{x},y))},
    \label{nu_expression}
\end{equation}
the expectancy of the radius of robustness becomes $\frac{\alpha}{\|\beta\|_2}$ where $\alpha$ is the test accuracy of the classifier on $\mathcal{D}$. In the case of SVMs, the quantity $\frac{1}{\|\beta\|_2}$ happens to be the margin between the classes, a notion obviously closely related to robustness. This motivates the following general definition:

\begin{definition}\label{def:robusnessours}
We call difficulty-aware robustness of a classifier $f$ over a dataset $\mathcal{D} = \{(\mathbf{x}_i,y_i)_{1\leq i \leq |\mathcal{D}|}\}$ the quantity:
$$\mathbb{E}_\nu\left[\left(r(\mathbf{x}_i,y_i)\right)_{1\leq i \leq |\mathcal{D}|}\right].$$
\end{definition}
where $\nu$ is a probability distribution on $\mathcal{D}$. It is pretty easy to check that irrespective of the probability distribution, the score satisfies properties (I) and (II) stated in section \ref{golden_score_section}. \par
In this article, our major statement is the following:
for any deep neural network using the cross-entropy loss $\ell$, $r(\mathbf{x},y)$ increases approximately linearly with $g(\ell(\mathbf{x},y))$, hence the score
\begin{equation}
\label{Rnu}
    R_{\nu}:=\mathbb{E}_\nu\left[\left(r(\mathbf{x}_i,y_i)\right)_{1\leq i \leq |\mathcal{D}|}\right] = \frac{1}{|\mathcal{D}|}\sum_{i=1}^{|\mathcal{D}|}\frac{r(\mathbf{x}_i,y_i)}{g(\ell(\mathbf{x}_i,y_i))}
\end{equation}
becomes a statistical quantity aiming at evaluating the slope of the line $(g(\ell(\mathbf{x},y)), r(\mathbf{x},y))$. Intuitively, the greater this slope is, the faster the radius of robustness increases as we move away from the class boundary to the center of the class. Implicitly, we set $\nu$ as in (\ref{nu_expression}).\par
Corollary \ref{corollary_binomial_regression} guarantees that this statement is completely true for binomial logistic regression and Proposition \ref{proposition_logistic} tells us it is pretty accurate for multinomial logistic regression. In the experiments section, we shall empirically confirm this statement for complex architectures and datasets, and see how this score predicts the quality of generalization in controlled synthetic cases and real-life datasets. Just before that, let us study how robustness can be seen through a probabilistic viewpoint.

\subsection{On the difficulty of estimating the radius of robustness}

A major problem with the proposed definition is that it requires to compute the radius of robustness of samples for a given classifier, which is expected to be hard in practice. Moreover, it is very different to be robust in every possible direction of space but a few, or to be robust in half the possible directions. To account for these notions, we also propose to incorporate a trade-off between the (probabilistic) radius $\varepsilon$ we consider and the probability $\alpha$ that the decision of the considered classifier is wrong.

\begin{definition}\label{def:robust_classi_our}
Given  $\alpha\in [0,1]$, we define the distribution-specific trade-off of robustness of a classifier $f$ for a random variable $Z$ and around a sample $\mathbf{x}$ of class $y$ as the largest $\varepsilon\geq 0$ such that: 
\begin{equation}\label{epsi}
  \P\left[  \ f(\mathbf{x}+\varepsilon Z) \neq y \right] \leq \alpha.  
\end{equation}
\end{definition}

In the remaining of this work, we often consider the isotropic trade-off of robustness, which is the distribution-specific trade-off obtained when $Z$ is a random variable uniformly distributed on the unit ball centered on $0$.

Now let us exhibit a bound on this isotropic trade-off of robustness. We consider here the $L^\infty$-norm, so that we can use the well known bound by Hoeffding~\cite{Hoeffding}:

\begin{theorem}
Let $X_1, ..., X_n$ be independent random variables such that $X_i$ takes values in the interval $[a_i, b_i]$. 
Set
$\overline {X}:=\frac {1}{n}(X_{1}+\cdots +X_{n})$.
Then:
\begin{equation}\label{hoeff_bound}
\P \left({\overline {X}}-\E \left[\overline {X}\right]\geq t\right)
\leq \exp\left(\frac{-2n^2t^{2}}{\sum_{i=1}^n (b_i-a_i)^2}\right)
\end{equation}
where $t\geq 0$.
\end{theorem}

We apply this bound in the following way: in the situation of binomial regression of parameters $(\beta, \beta^0)\in\mathbb{R}^d\times\mathbb{R}$, consider a given point $\mathbf{x}$ with $f(\mathbf{x})=1$, i.e.~$\beta_0 + \beta\cdot \mathbf{x} >0$.
Consider a noise variable $Z$ that is uniformly distributed on the $L^\infty$ unit ball, which means that the coordinates $Z_i$ of $Z$ are i.i.d. random variables with uniform distribution 
on $[-1,1]$. Then:
\begin{equation*}
  \P\left[  \ f(\mathbf{x}+\varepsilon Z) \neq y \right] = \P\left[\varepsilon \beta \cdot Z < -( \beta^0 + \beta\cdot \mathbf{x})\right]
=  \P\left[\beta\cdot Z > \frac{\beta_0 + \beta\cdot \mathbf{x}}{\varepsilon}\right]
\end{equation*}
by the symmetry of $Z$. Hence:
$$
\P[  \ f(\mathbf{x}+\varepsilon Z) \neq y ] 
=
\P\left[\sum_i \beta_i Z_i > \frac{\beta_0 + \beta\cdot \mathbf{x}}{\varepsilon}\right].
$$
To the right hand side we may apply Hoeffdings's inequality. Note that $X_i:=\beta_i Z_i$ is symmetrically distributed around the origin, with
$|X_i| \le \beta_i$ and $\E[X_i]:= \E[Z_i] =0$. Thus, using \eqref{hoeff_bound}:
\begin{eqnarray*}
\P[  \ f(\mathbf{x}+\varepsilon Z) \neq y ] 
\le \P\left[\frac 1 d \sum_i \beta_i Z_i > \frac{\beta_0 + \beta\cdot \mathbf{x}}{d \varepsilon}\right]\le
\exp\left( -\frac{ (\beta_0 + \beta\cdot \mathbf{x})^2}{2 \varepsilon^2 \sum_{i=1}^d \beta_i^2}\right) = \exp\left(-\frac{r^2}{2\varepsilon^2}\right),
\end{eqnarray*}
where $r$ is the radius of robustness of $\mathbf{x}$ for the $L^2$ norm. Let us choose $\varepsilon$ such that
\begin{equation}
    \alpha = \exp\left(-\frac{r^2}{2\varepsilon^2}\right), \quad \text{i.e.} \quad \varepsilon = \frac{r}{\sqrt{-2\log\left(\alpha\right)}} .
    \label{inequation_alpha}
\end{equation}
Interestingly, Equation~(\ref{inequation_alpha}) shows that it holds a threshold phenomenon: for a small value of $\alpha$, $\varepsilon$ vanishes quickly while for $\alpha$ closer to 1, $\varepsilon$ explodes.

For a fixed (small) value of $\alpha$, Equation (\ref{inequation_alpha}) indicates that if we can estimate the optimal value of  $\varepsilon$ satisfying (\ref{epsi}), this gives us an upper bound for the radius of robustness $r$. However, in practice, $\varepsilon$ is not easier to estimate than $r$. In the experiments, we directly estimate the values of $r$ by  simulations.

\section{Experiments}

In the following experiments, we work with the $L^2$ norm, unless mentioned otherwise.

\subsection{Synthetic data}

We first want to challenge the relevancy of our definition of robustness in Equation~(\ref{Rnu}) on completely controlled data and in the simple setup of logistic regression. To do so, we sample points from two two-dimensional isotropic Gaussian distributions of same standard deviation and different mean and call this dataset $\mathcal{D}$. We know the expression of the asymptotically best classifier, which is the median of the means of the two Gaussian distributions. We call this classifier the baseline and denote it by $(\beta^*, \beta^{0*})$. Now, we define a distance on linear classifiers. Let $(\beta_1, \beta^0_1)$ and $(\beta_2, \beta^0_2)$ defining two linear classifiers, then the distance between these two is defined as: $$d\left((\beta_1, \beta^0_1), (\beta_2, \beta^0_2)\right) := \|\beta_2-\beta_1\|_2 + |\beta^0_2- \beta^0_1|.$$

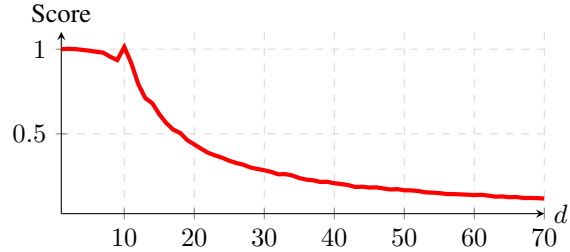
\begin{figure}[H]
    \centering
    \begin{tikzpicture}
	\begin{axis}[
	axis x line=middle,
	axis y line=middle,
	enlarge y limits=true,
	width=8cm, height=4cm,     
	grid = major,
	grid style={dashed, gray!30},
	ylabel=Score,
	xlabel=$d$,
every axis x label/.style={at={(current axis.right of origin)},anchor=west},
every axis y label/.style={at={(current axis.north west)},above=0mm},
	every axis plot/.append style={ultra thick}
	],        
	\addplot[color=red] table [x=a, y=b, col sep=comma] {data/dscores_ours.csv};
    \end{axis}
    \end{tikzpicture}
    \caption{Evolution of the score $R_{\nu}$ with respect to the distance to the base classifier. The score has been normalized so that it is equal to 1 at $d=0$.}
    \label{scores_distance}
\end{figure}

In this setup, we study the evolution of the proposed score $R_{\nu}(f_{\beta}, \mathcal{D})$ as a function of the distance $d((\beta^*, \beta^{0*}), (\beta, \beta^0))$, that is the distance of the classifier $(\beta, \beta^0)$ to the baseline. Namely, for each distance $d$, we randomly pick $n$ classifiers $(\beta, \beta^0)$ of distance $d$ to the baseline and we average their scores $R_{\nu}$.
The result of our experiment is depicted on figure (\ref{scores_distance}).

We observe some fluctuations around 1 at the beginning, which is not so surprising as $(\beta^*, \beta^{0*})$ is only \emph{asymptotically} best. Apart from that, the score $R_{\nu}$ has the awaited behaviour as it decreases smoothly as we go away from the asymptotically optimal classifier.

\subsection{Correlation between Loss and Robustness}
In the following, we investigate further for several architectures and real-life data the link between loss and radius of robustness. 

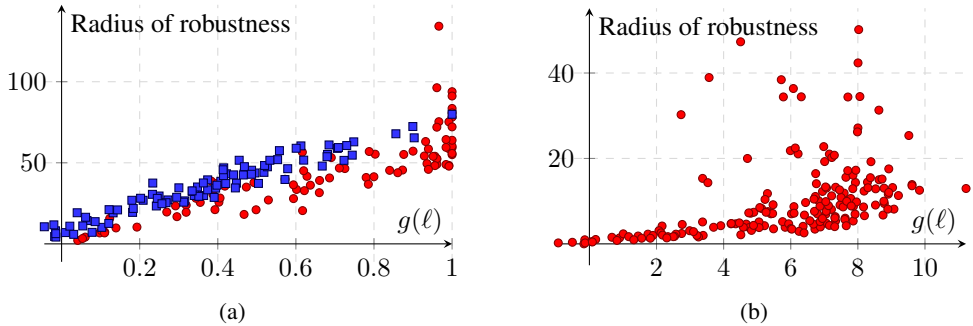
\begin{figure}[H]\centering
    \begin{subfigure}[t]{0.49\textwidth} \centering
		
	\begin{tikzpicture}
	\begin{axis}[
	axis x line=middle,
	axis y line=middle,
	enlarge y limits=true,
	width=7cm, height=5cm,     
	grid = major,
	grid style={dashed, gray!30},
	ylabel=Radius of robustness,
	xlabel=$g(\ell)$,
	legend style={at={(0.1,-0.1)}, anchor=north}
	]        
\addplot[only marks, mark options={draw=red!50!black, fill=red,}, mark size=1.5pt] table [x=a, y=b, col sep=comma] {data/regrllenet5100.csv};
	\addplot[only marks, mark options={mark=square*, fill=blue!80!white, draw=blue!30!black},  mark size=1.5pt] table [x=a, y=b, col sep=comma] {data/regrllogistic100.csv};
    \end{axis}
    \end{tikzpicture}
	\subcaption{}
	\label{rl_mnist}
	\end{subfigure}
	\begin{subfigure}[t]{0.49\textwidth} \centering
    \begin{tikzpicture}
	\begin{axis}[
	axis x line=middle,
	axis y line=middle,
	enlarge y limits=true,
	width=7cm, height=5cm,     
	grid = major,
	grid style={dashed, gray!30},
	ylabel=Radius of robustness,
	xlabel=$g(\ell)$,
	legend style={at={(0.1,-0.1)}, anchor=north}
	]        
	\addplot[only marks, mark options={draw=red!50!black, fill=red,},  mark size=1.5pt] table [x=a, y=b, col sep=comma] {data/rlresnet200.csv};
    \end{axis}
    \end{tikzpicture}
	\subcaption{}
	\label{rl_cifar}
	\end{subfigure}
	\label{mnist}
	\caption{Figure (\ref{rl_mnist}) shows points of the MNIST dataset scattered according to $g(\text{loss})$ and their radius of robustness. Blue and red dots respectively correspond to a logistic regression and to LeNet5. Figure (\ref{rl_cifar}) is the same type of plot for a ResNet18 trained on CIFAR-10 without data augmentation.}
\end{figure}

We train a multinomial logistic regression classifier on the MNIST dataset and, for 100 points in the validation set, plot their loss and their radii of robustness. They are depicted as blue dots on Figure (\ref{rl_mnist}). To compute this radius we proceed by dichotomy: at each step, we try in at most 5000 random directions of fixed norm to find a deformation that changes the class of the data point. If one is found we try with half the radius. We stop when the precision on the radius attains approximately $0.5$. 

In the case of the logistic regression, a linear regression gives us that, with a coefficient of determination of 0.93, the radius of robustness is a linear function of $g(\text{loss})$. We have the same kind of behaviour for LeNet5 and for a ResNet on CIFAR-10: except for some outliers, the radius still grows linearly with $g(\text{loss})$. Thus, we empirically verified that this behaviour holds in the case of a multinomial logistic regression, a CNN like LeNet5 and a Residual Neural Network trained with real data.\par

\subsection{A dataset-independent score}

For a logistic regression, the score $R_{\nu}$ is designed to be completely independent of the dataset samples it is evaluated upon. Let us stress if this property holds for complex neural networks, like LeNet5 trained on MNIST and a ResNet18 trained on CIFAR-10. We first randomly pick 100 samples from the validation sets. Then, we order them according to their loss: this is a pretty legitimate way of evaluating their difficulty. We split them in half to have in one hand the 50 easiest samples and in the other hand the 50 hardest ones. Finally we compute the mean score $R_m$ and the proposed score $R_{\nu}$ on both subsets and study their difference. Results are presented in Table~\ref{data_independant_score_table}. The relative variation (rel. var.) is defined as $|R^e-R^h|/(R^e+R^h)$ where $R^e$ and $R^h$ are respectively the scores on the easy and hard subsets.

\begin{table}[H]
\caption{$R_m$ and $R_{\nu}$ scores on easy and hard subsets of two datasets, MNIST and CIFAR-10, respectively computed for LeNet5 and a ResNet18.} 
\centering
	\begin{tabular}{@{}rrrrcrrrcrrr@{}}\toprule
		&\multicolumn{3}{c}{$R_m$} & \phantom{abc}& \multicolumn{3}{c}{$R_{\nu}$}\\\cmidrule{2-4} \cmidrule{6-8}& easy & hard & rel. var. && easy & hard & rel. var.\\
		\midrule
		LeNet5 & 55.2 & 24.6 & 0.38 && 1.63 & 1.91 & 0.08\\
		ResNet18 & 11.4& 4.5& 0.44&& 0.68& 0.54& 0.12\\
		\bottomrule
	\end{tabular}
\label{data_independant_score_table}
\end{table}

We notice that, for both networks, the relative variation between the mean scores on the easy and hard subsets is around 4 times the one of the score $R_{\nu}$. This shows empirically that the $R_{\nu}$ is less dependent of the dataset that $R_m$ is and therefore that it goes towards verifying property (III).

\begin{remark}
We also notice that in the case of LeNet5, the score $R_{\nu}$ is higher on the hard subset than on the easy one. This may indeed happen as this score measures the slope of the line $(g(\ell(\mathbf{x},y)), r(\mathbf{x},y))$ and not directly the radii.
\end{remark}

\subsection{Influence of adversarial training}

Next, we study the effect of increasing the robustness of a classifier. We choose to consider the work in \cite{madry2017towards}, where Madry and al. studied the effect of adversarial training on the MNIST and CIFAR-10 datasets. They released a GitHub repository \url{https://github.com/MadryLab/CIFAR-10_challenge.git} from which we downloaded pre-trained networks on CIFAR-10, both with and without adversarial training. Using this repository, we computed both the mean score of robustness $R_m$ and the difficulty-aware score $R_{\nu}$ for the two networks. In Table~\ref{adv_training_scores_table}, we present the results computed using 60 samples:

\begin{table}[H]
\centering
\caption{Scores of robustness on naturally and adversarially trained networks on CIFAR-10.}
\begin{tabular}{rrccrrrr}\toprule
		& $R_m$ && $R_{\nu}$ \\
		\midrule
		Naturally trained network & 2.4 && 0.25\\
		Adversarially trained network & 7.1&& 1.49\\
		\bottomrule
	\end{tabular}
\label{adv_training_scores_table}
\end{table}
We notice for both scores an increase. This supports the fact that these scores do measure well the gain in robustness of adversarially trained networks for complex datasets.

\section{Conclusion}

In this paper, we pointed out the limits of using the mean radius to measure the robustness of a classifier. We showed that it is expected that some samples yield larger radii than others, depending on their difficulty and typicality. As such, we designed a simple score of robustness that accounts for this variation. We proved this score is theoretically independent of the choice of the samples for the simple case of a binomial logistic regression. We also derived multiple experiments with various datasets and neural network architectures to demonstrate that it still provides an interesting measure of robustness for more complex settings. Finally, the computational cost of this score is the same as the classical mean-case $R_m$ and worst-case $R_w$ as it is only needed to compute the losses of the samples on top of the radii, which consists in a negligible additional cost. In future work, it would be interesting to investigate whether sampling training inputs depending on their distance to the class boundary could result in more robust trained classifiers.

\section{Broader impact}

Adversarial attacks can prove to be life-threatening: in \cite{eykholt2018robust}, the authors showed that it is possible to fool state-of-the-art classifiers by simply putting small stickers on stop signs. The classifiers then believe they are looking at a speed-limit sign, which can of course lead to a very dangerous behaviour from an autonomous car driven by a neural network. Also, \cite{hendrycks2019benchmarking} showed that real-life perturbations like rain or fog can considerably damage the performance of a classifier. It is therefore of first importance to measure and eventually increase the robustness of Deep Learning architectures. In this context, our work goes towards this goal by giving a more intrinsic and accurate measure of robustness. \par
Furthermore, robustness is intrinsically linked to explainability \cite{tsipras2018robustness}, the latter being a central issue in today's law regarding Artificial Intelligence. That is why increasing the robustness of Deep Learning architectures can only facilitate the wide use of this technology in the near future.

\bibliographystyle{abbrv}
\bibliography{biblio}

\end{document}


\maketitle

\section{Proof of proposition (3.2)}

We give here the proof of proposition (3.2) which holds in the context of multinomial logistic regression:

\begin{customthm}{3.2}
For every well-classified input $\mathbf{x}$ belonging to class $k$, there exists $m\not=k$ such that
	\begin{equation}
		\frac{-1}{\|\beta_m-\beta_k\|}\log\bigg(\frac{1}{K-1}(\exp\ell(\mathbf{x},k)-1)\bigg) \leq r(\mathbf{x},k) \leq \frac{-1}{\|\beta_m-\beta_k\|}\log\big(\exp\ell(\mathbf{x},k)-1\big),
    \end{equation}
where $r$ is defined for $q=2$, i.e. the euclidean norm.
\end{customthm}

\begin{proof}
	Because we suppose that the logistic regression classifies well $\mathbf{x}$, we have
	\begin{equation}
		\exp(\beta_k\cdot\mathbf{\mathbf{x}}+\beta_k^0) \geq \exp(\beta_j\cdot\mathbf{\mathbf{x}}+\beta_j^0), \quad \forall j\in\{1, \dots, K\}.
	\end{equation}
	The radius of robustness is defined as the smallest $r\geq0$ such that 
	\begin{multline}
		\exists \mathbf{n}\in\mathbb{R}^d, \|\mathbf{n}\|=r ~|~ \exists m\in\{1, \dots, K\} ~\text{such that}~ \\
		 \exp(\beta_k\cdot(\mathbf{x+n})+\beta_k^0) = \exp(\beta_m\cdot(\mathbf{x+n})+\beta_m^0),
	\end{multline}
	i.e.
	\begin{equation}
	\label{hyperplane}
		(\beta_k-\beta_m)\cdot(\mathbf{x+n})+(\beta_k^0-\beta_m^0) = 0.
	\end{equation}
	Hence the radius of robustness is the distance of $\mathbf{x}$ from the hyperplane defined by equation (\ref{hyperplane}), i.e.
	\begin{equation}
		r(\mathbf{x},k) = \frac{(\beta_k-\beta_m)\cdot\mathbf{x}+(\beta_k^0-\beta_m^0)}{\|\beta_k-\beta_m\|}.
	\end{equation}
	Now, we have
	\begin{align}
	\ell(\mathbf{x}, k) &= -\log \Bigg(\frac{\exp(\beta_k\cdot\mathbf{x}+\beta_k^0)}{\sum_{j=1}^K\exp(\beta_j\cdot\mathbf{x}+\beta_j^0)}\Bigg) \\
	&= -\log \Bigg(\frac{1}{ \sum_{j=1}^K\exp\big((\beta_j-\beta_k)\cdot\mathbf{x}+(\beta_j^0-\beta_k^0)\big)}\Bigg) \\
	&= \log \Bigg(\sum_{j=1}^K\exp\big((\beta_j-\beta_k)\cdot\mathbf{x}+(\beta_j^0-\beta_k^0)\big)\Bigg) \\
	&= \log \Bigg(1+\exp(-\|\beta_m-\beta_k\|\cdot r(\mathbf{x},k))\label{main_ineq}\\&+\sum_{j\not\in\{k,m\}}\exp\big((\beta_j-\beta_k)\cdot\mathbf{x}+(\beta_j^0-\beta_k^0)\big)\Bigg). \nonumber 
	\end{align}
	Hence, 
	\begin{equation}
		l(\mathbf{x},k)\geq \log\Big(1+\exp\big(-\|\beta_m-\beta_k\|\cdot r(\mathbf{x},k)\big)\Big).
	\end{equation}
	To get the theorem's second inequality, it suffices to invert this expression. \par
	Now, $\forall j\not\in\{k,m\}$,
	\begin{align}
		\exp\big((\beta_j-\beta_k)\cdot\mathbf{x}+(\beta_j^0-\beta_k^0)\big) &= \exp\big((\beta_j-\beta_m)\cdot\mathbf{x}+(\beta_j^0-\beta_m^0)\big)\\&\times\exp\big((\beta_m-\beta_k)\cdot\mathbf{x}+(\beta_m^0-\beta_k^0)\big)\nonumber\\
		&\leq 1\times \exp(-\|\beta_m-\beta_k\|\cdot r(\mathbf{x},k)).
	\end{align}
	Hence from (\ref{main_ineq}) we get 
	\begin{equation}
		l(\mathbf{x},k)\leq \log\Big(1 + (K-1)\exp\big(-\|\beta_m-\beta_k\|\cdot r(\mathbf{x},k)\big)\Big).
	\end{equation}
	To get the theorem's first inequality it suffices again to invert this expression.
\end{proof}

\section{Further experiments about the dataset quality}

In this section, we challenge further our idea that every samples have different qualities. More precisely, we see in the following experiments that some samples, without being great in number, worsen a lot the robustness of a classifier.

\subsection{Heat map on CIFAR-10}
\label{Hmap_section}
\begin{figure}[H]\centering
	\begin{subfigure}[t]{0.49\textwidth} \centering
	\begin{tikzpicture}
    \begin{axis}[
    zmin=-10,zmax=10,
    view={0}{90},
    width=6cm, height=6cm,
    ylabel=Number of samples kept $d$,
	xlabel=Intensity of the noise,
    colormap/viridis,
    colorbar,
    y dir=reverse
    ]
    \addplot3 [
    surf,
    ] table [
    x index=0,
    y index=1,
    z expr=\thisrowno{2},
    ] {data/Hmap.csv};
    \end{axis}
    \end{tikzpicture}
	\subcaption{}
	\label{Hmap}
	\end{subfigure}
	\begin{subfigure}[t]{0.49\textwidth} \centering
		
	    \begin{tikzpicture}
	\begin{axis}[
	axis x line=middle,
	axis y line=middle,
	enlarge y limits=true,
	width=7cm, height=7cm,     
	grid = major,
	grid style={dashed, gray!30},
	ylabel=MCE,
	xlabel style={text width=2.5cm}, xlabel=Length of\\the dataset,
	legend style={at={(0.1,-0.1)}, anchor=north},
	x label style={at={(axis description cs:1.09,+0.26)},anchor=north},
	]        
	\addplot[color=red, ultra thick] table [x=a, y=b, col sep=comma] {data/dlengthMREresnet500.csv};
    \end{axis}
    \end{tikzpicture}
	\subcaption{}
	\label{RatioMCE}
	\end{subfigure}
	\caption{Figure (\ref{Hmap}) shows the accuracy heat map, which details can be found at the beginning of section \ref{Hmap_section}. Figure (\ref{RatioMCE}) shows the evolution of the Mean Corruption Error as a function of the quality of the dataset.}
\end{figure}
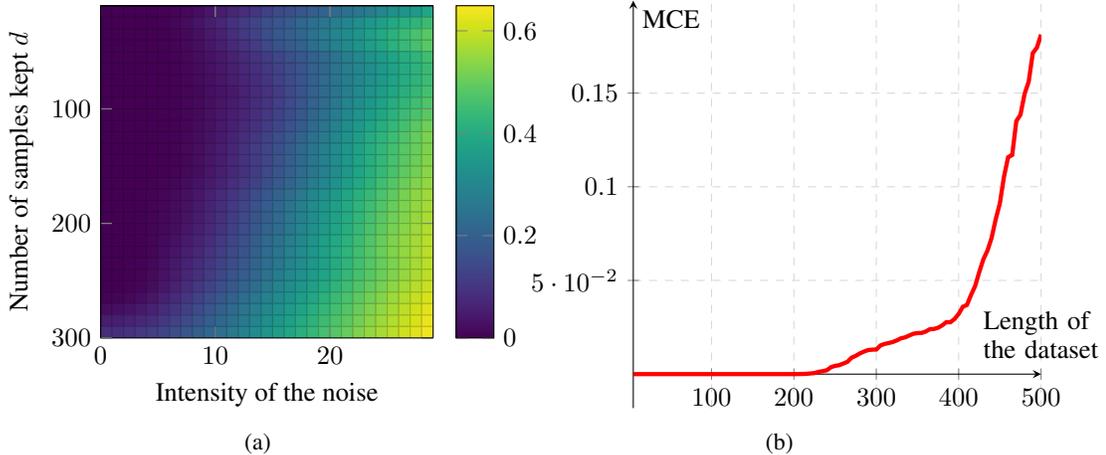

We work here with a randomly picked subset of size 300 of the validation set of CIFAR-10. Our classifier is a ResNet18 trained on CIFAR-10 without data augmentation. We now plot a heat map with the heat being the accuracy error of this network on the chosen subset, see Figure (\ref{Hmap}). The latter subset is distorted with a random Gaussian noise of euclidean norm $r$ ($x$-axis). Furthermore, we reduce the subset size from 300 to $d$ ($y$-axis) by keeping the $d$ samples (not distorted) with the lowest loss. In order to smooth the map, we make 50 simulations for each point $(x,y)$. \par
Thank to this heat map, we can study at the same time the evolution of the accuracy error regarding to
\begin{itemize}
    \item the intensity of the noise applied to the inputs ($x$-axis)
    \item the quality of the dataset ($y$-axis)
\end{itemize}

We see that if we keep all the data set, i.e. for the line corresponding to $y=300$, there is no minimum radius of robustness. However, if we begin to remove pathological samples, that is with high loss, we observe the existence of a radius of protection (the dark blue area). 

\subsection{Mean Corruption Error on CIFAR-10}

In \cite{hendrycks2019benchmarking}, Hendrycks and al. suggest several kind of deformations for the CIFAR-10 and ImageNet datasets that can happen in real-life, like rain, fog, JPEG compression etc. For a ResNet18 trained without data augmentation on CIFAR-10, we compute the errors corresponding to each deformation. We call their mean the Mean Corruption Error (MCE). We plot this quantity for different size of the data set, again selected according to the loss. We again perform an averaging procedure: for each length $\ell$ of the data set, we select several random subset (of size 300) of the total data set (of size 10000). For each subset, we select the $\ell$ samples with the lowest loss and compute with them the MCE. Finally, we averaged the obtained values. The result is depicted in Figure (\ref{RatioMCE}). \par
We observe three distinct trends. First, when the data set is only composed of easy (i.e. with low loss) samples, the MCE is equal to 0. Then, around the middle size of the data set, the MCE begins to grow linearly. Finally, next to the four fifths, the MCE grows much faster and also linearly. This behaviour confirms the fact that a small number of samples, here one fifth, has a very big impact on the measured robustness of a classifier.

\bibliographystyle{abbrv}
\bibliography{biblio}